\documentclass[sigconf]{acmart}
\usepackage{tikz}
\usetikzlibrary{arrows.meta, positioning, shadows}

\definecolor{Slate} {HTML}{475569}
\definecolor{Indigo}{HTML}{4F46E5}
\definecolor{Amber} {HTML}{D97706}
\definecolor{Teal}  {HTML}{0D9488}
\usepackage{xcolor}
\usepackage{soul}

\AtBeginDocument{%
  }

\setcopyright{none}
\copyrightyear{2026}
\acmYear{2026}
\acmDOI{} 
\acmISBN{} 

\acmConference[AI4IR '26]{The 2026 Workshop on AI Agent for Information Retrieval (Agent4IR), co-located with ACM SIGKDD 2026}{August 09--13, 2026}{Jeju, Korea}
\acmBooktitle{The 2026 Workshop on AI Agent for Information Retrieval (AI4IR '26), co-located with ACM SIGKDD 2026, August 09--13, 2026, Jeju, Korea}

\begin{document}

\title{PromptPack: Scaling LLM Annotation Agents\\ for Online Recommendation}

\author{Sebastian Koralewski}
\affiliation{%
  \institution{Teads Inc.}
  \city{New York}
  \state{New York}
  \country{USA}
}

\author{Merwan Barlier}
\affiliation{%
  \institution{Teads Inc.}
  \city{New York}
  \state{New York}
  \country{USA}
}

\author{Yulia Stolin}
\affiliation{%
  \institution{Teads Inc.}
  \city{New York}
  \state{New York}
  \country{USA}
}

\author{Bla\v{z} \v{S}krlj}
\affiliation{%
  \institution{Teads Inc.}
  \city{New York}
  \state{New York}
  \country{USA}
}

\renewcommand{\shortauthors}{Koralewski et al.}

\begin{abstract}
  Online recommendation platforms increasingly use Large Language Models (LLMs) to extract structured features from ad creatives. While deploying a single-call LLM annotation agent yields significant Click-Through Rate (CTR) improvements in our live production environment, per-creative prompting is prohibitively expensive to scale. The redundant system instructions sent in every request account for 94\% of billed input tokens. To break this cost bottleneck, we introduce PromptPack, a scalable, high-throughput LLM annotation agent. PromptPack achieves this scale via in-context batching, combining a shared system prompt, a strict XML structural envelope, and an output correction layer to ensure deterministic, pipeline-ready feature extraction across multiple creatives simultaneously. We evaluate PromptPack via an offline retrieval benchmark using a downstream logistic-regression ranker. To deeply profile the agent's behavior, we measure AUC and introduce Volume-Weighted Absolute Lift (VWAL), a novel metric capturing the signal quality of the generated features. Compared to our live, unbatched production baseline, PromptPack at batch size 20 cuts our LLM costs by 89\% and accelerates throughput by 2.5x while fully preserving AUC.
\end{abstract}

\ccsdesc[500]{Information systems~Recommender systems}
\ccsdesc[500]{Computing methodologies~Natural language processing}
\ccsdesc[300]{Information systems~Online advertising}
\ccsdesc[500]{Information systems~Information retrieval}
\keywords{Recommender Systems, Information Retrieval, Large Language Models, In-Context Batching, Feature Extraction, Click-Through Rate Prediction, Inference Efficiency}

\maketitle
\section{Introduction}
\label{sec:intro}

Click-through rate (CTR) prediction in online advertising is a foundational information-retrieval problem \cite{broder2008computational,dave2014computational}. To match a user with the most relevant ad, platforms retrieve and rank candidates using CTR models that rely heavily on creative-side features such as semantic content and target audience details. Extracting high-quality features from ad creatives with Large Language Models (LLMs) significantly boosts this retrieval accuracy. Specifically, frontier LLMs can now generate these rich, multi-label tags directly from short ad titles and metadata \cite{liu-etal-2025-real,10.1145/3711896.3737242}.

To exploit these capabilities, we previously deployed an LLM annotation agent on \texttt{gpt-4.1-nano} in our live production environment~\cite{agent0}. By issuing a single-call LLM request and the full feature-taxonomy prompt per individual ad, this agent successfully delivered a $+0.03\%$ relative information gain (RIG) lift on our production CTR scorer, demonstrating the immense value of LLM-derived features. However, our feature taxonomy is substantial (roughly $8{,}500$ tokens), meaning that $94\%$ of the billed input tokens in every request repeat static content that never changes. This massive overhead creates a severe cost bottleneck, squeezing margins and preventing us from widening the rollout to include additional retrieval features.

The standard approach to eliminating such overhead is in-context batching (or batch prompting) \cite{cheng2023batchprompting}, an inference technique that groups multiple queries into a single prompt to amortize the system context cost. However, merging multiple independent queries into a single context window introduces severe technical difficulties. Because language models process sequences fluidly, naive batching frequently suffers from context bleeding (or semantic cross-talk), where highly weighted features or keywords from one ad creative inadvertently contaminate the feature extraction of an adjacent item in the batch. This loss of input isolation, compounded by severe positional bias, typically degrades downstream retrieval accuracy and prevents straightforward list-based merging in production systems.

To resolve this trade-off, we introduce PromptPack, a scalable, high-throughput LLM annotation agent architecture that successfully batches multiple inputs into a single LLM call while entirely mitigating context bleeding. PromptPack combines three core elements to achieve this: (1) a \textbf{Shared System Prompt} evaluated exactly \emph{once per batch} to eliminate token redundancy; (2) a strict \textbf{XML Structural Envelope} providing unambiguous item boundaries to enforce attention isolation and prevent semantic cross-talk; and (3) an \textbf{Output Correction Layer} that guarantees the LLM's best-effort JSON is deterministically converted into pipeline-ready feature rows.

While our cost-reduction objective centers on the \texttt{gpt-4.1-nano} production baseline, we benchmark PromptPack across a panel of four foundation models to serve two purposes:
\begin{enumerate}
    \item \textbf{Generalization Evidence:} To empirically verify if PromptPack's structural batching benefits extend robustly across different LLMs from different providers.
    \item \textbf{Forward-Looking Candidates:} To identify whether larger, faster frontier models yield downstream accuracy gains that justify their higher deployment costs.
\end{enumerate}

We evaluate PromptPack via an offline retrieval benchmark using a production-derived data set of $10{,}000$ creatives, leveraging a downstream logistic-regression ranker to measure predictive utility. To deeply profile the agent's behavior, we measure AUC and introduce Volume-Weighted Absolute Lift (VWAL), a diagnostic metric that isolates total feature signal volume from ranking accuracy. 

Our empirical results confirm our primary objective: at a batch size of 20, PromptPack cuts token costs by $89\%$ and accelerates throughput by $2.5\times$ while successfully matching the baseline production model's prediction quality. This efficiency win generalizes across all tested architectures. Furthermore, the benchmark identifies \texttt{claude-haiku-4.5} and \texttt{gemini-2.5-flash} as producing the highest absolute AUCs, establishing them as strong candidates for future live production trials.

This work is structured as follows: Section~\ref{sec:method} details the PromptPack architecture. Section~\ref{sec:vwal} introduces the VWAL diagnostic metric. Section~\ref{sec:exp} outlines the evaluation setup and baselines. Section~\ref{sec:results} reports our empirical results, and Section~\ref{sec:conclusion} concludes the paper.
\section{Related Work}
\label{sec:related_work}
The financial and computational bottlenecks associated with Large Language Model (LLM) inference have driven a surge of research into cost-reduction and throughput-optimization techniques, which broadly fall into prompt-level batching strategies, system-level infrastructure modifications, and model routing. Batch prompting was initially formalized by Cheng et al.~\cite{cheng2023batchprompting}, who demonstrated that concatenating multiple inputs into a single prompt significantly reduces token usage by sharing system instructions and few-shot demonstrations. However, because naive batching often suffers from positional bias and degraded output quality, subsequent frameworks introduced complex heuristic management. For instance, BatchPrompt~\cite{lin2024batchprompt} employs Batch Permutation and Ensembling (BPE) alongside Self-reflection-guided Early Stopping (SEAS) to combat position bias via majority voting, albeit at the cost of requiring multiple inference passes. Other works focus heavily on the semantic grouping of prompts prior to inference. CliqueParcel~\cite{cliqueparcel2024} utilizes a pre-model to classify and group prompts into specific ``clique domains,'' ensuring only conceptually similar tasks are batched together. Similarly, frameworks like Batcher and Optimized Batch Prompting (OBP)~\cite{ji2025optimized} frame batching as a constrained optimization problem, relying on external embedding models to calculate pairwise affinities between queries and demonstrations to actively cluster similar questions. Beyond prompt manipulation, several studies optimize the underlying inference engines or model selection pipelines. Systems like BatchLLM~\cite{zheng2024batchllm} introduce horizontal fused prefix-shared attention and decode-first scheduling, achieving massive throughput gains by modifying the Key-Value (KV) cache memory management at the GPU kernel level. Alternatively, cascading frameworks like FrugalGPT~\cite{chen2024frugalgpt} avoid sequence batching altogether, instead training a generation judger to dynamically route queries to the smallest, cheapest model capable of answering them. While all of the aforementioned techniques have demonstrated highly promising results in academic benchmarks, they often introduce severe operational friction in fast-moving production environments. First, many of these frameworks rely on cost-driven preprocessing overhead. For example, utilizing external LLMs or embedding models to calculate query affinity and cluster inputs~\cite{ji2025optimized,cliqueparcel2024} adds latency and pipeline complexity. Second, they often depend heavily on creative input and restrictive heuristics, requiring queries to be strictly batched into highly similar semantic groups to achieve acceptable accuracy. Furthermore, system-level optimizations~\cite{zheng2024batchllm} require practitioners to host local LLMs, incur high operating costs, and perform deep, brittle engineering modifications to attention kernels and CUDA memory management. In contrast, we strive for a more straightforward, model-agnostic solution. Real-world industry applications—such as high-volume ad-click feature enrichment—demand a scalable approach that does not require expensive preprocessing, semantic clustering, or complex local infrastructure. Our work addresses this gap by proposing a lightweight, markup-guided sequence packing strategy that achieves substantial speedups and cost reductions directly via standard APIs, maximizing profit without sacrificing downstream predictive quality.
\section{The PromptPack Annotation Agent}
\label{sec:method}

\begin{figure}[t]
\centering
\begin{tikzpicture}[
  node distance=7mm and 14mm,
  every node/.style={font=\scriptsize},
  >={Latex[length=2.2mm,width=1.8mm]},
  box/.style={rectangle, draw, rounded corners=3pt, align=center,
              inner sep=4pt, minimum height=9mm, thick,
              drop shadow={shadow xshift=0.4pt, shadow yshift=-0.4pt,
                           opacity=0.15, fill=black}},
  data/.style   ={box, fill=gray!8,    draw=gray!55,    minimum width=24mm},
  agent/.style  ={box, fill=blue!7,    draw=blue!55,    minimum width=34mm,
                  minimum height=18mm},
  service/.style={box, fill=white,     draw=black!55,   minimum width=20mm,
                  densely dashed},
  model/.style  ={box, fill=orange!12, draw=orange!75!black, minimum width=24mm},
  output/.style ={box, fill=green!8,   draw=green!55!black,  minimum width=24mm},
  stage/.style  ={font=\scriptsize\itshape\color{black!55}},
  flow/.style   ={->, thick, draw=black!70},
  api/.style    ={->, thick, densely dashed, draw=blue!55!black},
]

\node[data] (creatives) {Ad creatives\\\textcolor{black!55}{\tiny(inventory)}};

\node[agent, below=of creatives]
  (agent) {\textbf{PromptPack agent}\\[1pt]
           \textcolor{black!70}{\tiny taxonomy prompt $\,\cdot\,$ XML structure
           $\,\cdot\,$ correction layer}};

\node[service, right=of agent] (llm)
  {LLM API\\\textcolor{black!55}{\tiny small/fast tier}};

\node[data, below=of agent] (feats)
  {Retrieval feature store\\\textcolor{black!55}{\tiny structured tag features}};

\node[model, below=of feats] (ctr) {CTR retrieval scorer};

\node[output, below=of ctr] (bid)
  {Ranked ad list\\\textcolor{black!55}{\tiny to downstream auction}};

\node[stage, left=2mm of creatives.west, anchor=east] {input};
\node[stage, left=2mm of agent.west,     anchor=east] {annotation};
\node[stage, left=2mm of feats.west,     anchor=east] {features};
\node[stage, left=2mm of ctr.west,       anchor=east] {scoring};
\node[stage, left=2mm of bid.west,       anchor=east] {output};

\draw[flow] (creatives) -- (agent);
\draw[flow] (agent)     -- (feats);
\draw[flow] (feats)     -- (ctr);
\draw[flow] (ctr)       -- (bid);

\draw[api] ([yshift= 2mm]agent.east) -- ([yshift= 2mm]llm.west)
  node[midway, above, font=\tiny, inner sep=1pt]{batch prompts};
\draw[api] ([yshift=-2mm]llm.west)   -- ([yshift=-2mm]agent.east)
  node[midway, below, font=\tiny, inner sep=1pt]{XML response};

\end{tikzpicture}
\caption{PromptPack annotation agent inside the ad-retrieval
pipeline. The agent translates raw creatives into the structured tag
features consumed by the CTR model. An external LLM is
queried in batches and its XML output passes through a correction
layer before reaching the feature store.}
\label{fig:arch}
\end{figure}
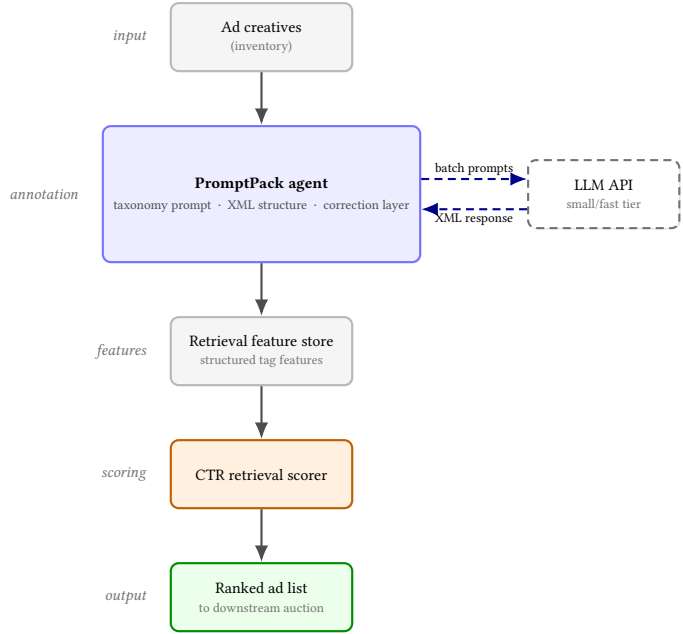

Motivated by the need for a straightforward, production-ready solution that avoids the operational friction and preprocessing overhead of prior methods, we introduce PromptPack. PromptPack is a lightweight, LLM-based annotation agent that sits inside the ad-retrieval pipeline (Fig.~\ref{fig:arch}) between the raw creative inventory and the retrieval system's feature store.

Conceptually, the agent has a single, highly scalable job: take a raw ad title in and return a structured, multi-tag feature record out, suitable for direct consumption by downstream models, such as click-through rate (CTR) prediction.

Rather than relying on expensive semantic clustering, external embedding models, or brittle architectural modifications, PromptPack achieves high-throughput batching entirely at the API level. To accomplish this, the agent's architecture is built upon three core, model-agnostic elements: a comprehensive taxonomy prompt, a robust XML structure to enable markup-guided sequence packing, and a lightweight correction layer to handle edge cases without needing complex model-routing networks.


The further considered setting is aligned with deployed production environment, yet not exact mirror due to required disclosure policies.
The agent's input is a short ad-creative title. Its output is a row of five feature columns containing
underscore-joined \texttt{tag;confidence} pairs. The five features
are \texttt{topic},  \texttt{sentiment}, \texttt{entity\_types},
\texttt{intent}, \texttt{style}. Each is multi-label, with example
vocabularies summarised in Table~\ref{tab:features}.

\begin{table}[ht!]
\centering
\caption{The five features the agent emits. Vocabularies are open (example-driven) and wire-format is strict.}
\label{tab:features}
\small
\begin{tabular}{|l|l|}
\hline
Feature & Example tags \\
\hline
\texttt{topic} & politics, finance, technology, science, \\
& health, sports, food-drink, travel, \dots \\
\texttt{entity\_types} & person, organization, location, product, \\
& event, monetary, date-time, numeric, \dots \\
\texttt{sentiment} & positive, negative, neutral, joy, anger, \\
& fear, anticipation, trust, disgust, \dots \\
\texttt{intent} & inform, instruct, recommend, persuade, \\
& warn, sell, entertain, question, \dots \\
\texttt{style} & formal, informal, technical, journalistic, \\
& sensational, listicle, conversational, \dots \\
\hline
\end{tabular}
\end{table}

The taxonomy is open, not closed. This implies agent can, on its own, define and extend new categories in alignment with data stream. The system prompt lists
example tags but the LLM is free to emit new tags when the input
warrants. Wire-format is enforced: lowercase ASCII with
hyphen-separated words, sorted by descending confidence,
deduplicated, capped at six tags per feature, floored at confidence
$0.10$ (lower confidences are not emitted. Tag becomes \texttt{none;0.2}). Temperature is zero throughout, so a given creative produces the same tag set across reruns.

\subsection{Feature Taxonomy Prompt}
\label{sec:method-prompts}
The first element of PromptPack is a thoughtfully constructed shared system prompt. In our production environment, this comprehensive prompt spans approximately 8,500 tokens. At this magnitude, processing ads sequentially incurs prohibitive latency and token costs. By prepending this massive taxonomy context once per call rather than once per ad, PromptPack aggressively amortizes the overhead across the entire batch.

Achieving a reliable, production-grade prompt of this size required extensive, iterative prompt engineering. When packing multiple items into a single inference call, LLMs are naturally prone to losing focus, mixing contexts between items, or hallucinating tags outside the desired vocabulary. To suppress these failure modes and guarantee a strict output schema, the prompt was heavily refined into a rigid, top-to-bottom structure comprising five sections:

(i) an introduction framing the LLM as a strict feature-extraction system; (ii) format rules specifying the exact wire format; (iii) a confidence rubric defining empirical bounds (e.g., what 0.9 vs. 0.7 vs. 0.3 mean concretely); (iv) detailed per-feature definitions equipped with example tag vocabularies and worked input/output pairs to ground the model's reasoning and (v) a strict batch-mode addendum dictating how the output must align to row identifiers and exactly what each per-item record must look like.


\subsection{XML Structure}
\label{sec:method-xml}
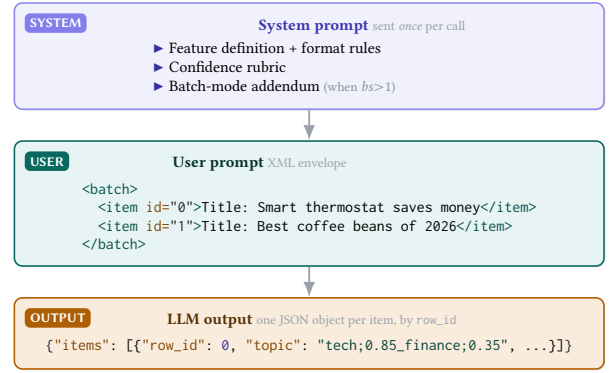
\begin{figure}[t]
\centering
\begin{tikzpicture}[
  every node/.style={font=\scriptsize},
  >={Latex[length=2mm, width=1.6mm]},
  layer/.style={rectangle, draw, rounded corners=3pt,
                minimum width=78mm, align=left, inner sep=6pt,
                line width=0.6pt,
                drop shadow={shadow xshift=0.5pt, shadow yshift=-0.5pt,
                             opacity=0.18, fill=black!60}},
  sys/.style ={layer, fill=Indigo!8,  draw=Indigo!65},
  usr/.style ={layer, fill=Teal!10,   draw=Teal!70!black},
  outl/.style={layer, fill=Amber!14,  draw=Amber!80!black},
  tag/.style={rectangle, rounded corners=2pt, inner sep=2pt,
              font=\tiny\sffamily\bfseries, text=white},
  meta/.style={font=\tiny, text=Slate!65},
  flow/.style={->, line width=0.7pt, draw=Slate!55},
]

\node[sys] (sys) {%
  \hspace*{14mm}\textbf{\textcolor{Indigo!75!black}{System prompt}}
  \hfill \textcolor{Slate!55}{\tiny sent \emph{once} per call}\\[2pt]
  \textcolor{Indigo!70!black}{$\blacktriangleright$}~Feature definition + format rules\\
  \textcolor{Indigo!70!black}{$\blacktriangleright$}~Confidence rubric\\
\textcolor{Indigo!70!black}{$\blacktriangleright$}~Batch-mode addendum {\tiny\textcolor{Slate!65}{(when $\mathit{bs}{>}1$)}}%
};
\node[tag, fill=Indigo!70, anchor=north west]
  at ([xshift=4pt, yshift=-4pt]sys.north west) {SYSTEM};

\node[usr, below=4mm of sys] (usr) {%
  \hspace*{12mm}\textbf{\textcolor{Teal!35!black}{User prompt}}
  \hfill \textcolor{Slate!55}{\tiny XML envelope}\\[3pt]
  {\ttfamily\textcolor{Teal!55!black}{<batch>}}\\
  {\ttfamily\quad\textcolor{Teal!55!black}{<item} \textcolor{Amber!60!black}{id=}\textcolor{Slate!40!black}{"0"}\textcolor{Teal!55!black}{>}Title: Smart thermostat saves money\textcolor{Teal!55!black}{</item>}}\\
  {\ttfamily\quad\textcolor{Teal!55!black}{<item} \textcolor{Amber!60!black}{id=}\textcolor{Slate!40!black}{"1"}\textcolor{Teal!55!black}{>}Title: Best coffee beans of 2026\textcolor{Teal!55!black}{</item>}}\\
  {\ttfamily\textcolor{Teal!55!black}{</batch>}}%
};
\node[tag, fill=Teal!70!black, anchor=north west]
  at ([xshift=4pt, yshift=-4pt]usr.north west) {USER};

\node[outl, below=4mm of usr] (outl) {%
  \hspace*{16mm}\textbf{\textcolor{Amber!40!black}{LLM output}}
  \hfill \textcolor{Slate!55}{\tiny one JSON object per item, by \texttt{row\_id}}\\[3pt]
  {\ttfamily\textcolor{Slate!50!black}\{\textcolor{Amber!60!black}{"items"}: [\{\textcolor{Amber!60!black}{"row\_id"}: \textcolor{Indigo!70!black}{0}, \textcolor{Amber!60!black}{"topic"}: \textcolor{Teal!45!black}{"tech;0.85\_finance;0.35"}, ...\}]\textcolor{Slate!50!black}\}}%
};
\node[tag, fill=Amber!80!black, anchor=north west]
  at ([xshift=4pt, yshift=-4pt]outl.north west) {OUTPUT};

\draw[flow] (sys) -- (usr);
\draw[flow] (usr) -- (outl);

\end{tikzpicture}
\caption{The three layers of a PromptPack call: a shared system
prompt, an XML-enveloped user prompt, and a JSON output aligned by
\texttt{row\_id}.}
\label{fig:prompt}
\end{figure}
Both academic work and practitioner guidance recommend XML formatting as an effective strategy for creating clear, unambiguous boundaries within complex prompts \cite{xmlprompting2025,anthropic2024xml}. To implement our lightweight, model-agnostic solution, we leverage this insight by wrapping each batched call's items in a strict XML envelope (Fig.~\ref{fig:prompt}). The batch is structured as a \texttt{<batch>} element containing one \texttt{<item id="N">...</item>} child per creative, with any XML special characters in the ad titles safely escaped. 

This XML envelope serves three distinct purposes. First, it provides critical delimiter robustness. Ad titles frequently contain commas, pipes, leading digits, or em-dashes that easily confuse simpler list formats. By enforcing explicit \texttt{<item>} tags, we create rigid, programmatic boundaries that are immune to these parsing errors.

Second, an \emph{explicit row-identifier contract}: the \texttt{id}
attribute is referenced in the system-prompt addendum as the field
the LLM must copy into each output record, so a reordered or
partial response can still be joined back. Third, a \emph{familiar
idiom}: every instruction-tuned LLM has seen XML many times in
training, so small fast-tier models stay aligned across many items
where they would otherwise drift.

The request additionally includes a one-line response-format
directive specifying valid JSON. We do not use a strict structured-output schema at decode time, because an earlier experiment showed that strict schemas suppressed AUC, likely by overly constraining the model's generative flexibility. Format compliance is enforced through the worked example in the system prompt plus
the correction layer below.

\subsection{Correction Layer}
\label{sec:method-correction}

The correction layer turns the LLM's best-effort JSON into a
deterministic, retrieval-pipeline-ready row and is backed by an extensive test suite. It runs as a sequence
of stages: a JSON parser with repair heuristics (closing dangling
braces at truncation points, normalising whitespace, repairing
malformed confidences, removing trailing commas); a tag-level
regex validator that filters out illegal spellings and back-fills
missing confidences; a row-identifier alignment check that drops
items with missing, duplicated, or out-of-range \texttt{row\_id};
tag-count and confidence-floor enforcement and a bounded per-item
retry pass that re-issues any creative the first pass failed to
enrich, at batch size one with an explicit ``this is a retry''
hint. Retry phases are capped at $5\%$ of the data set to avoid
unbounded cycles.



\section{Volume-Weighted Absolute Lift}
\label{sec:vwal}

We adopt downstream ROC-AUC as our primary evaluation metric because it directly reflects the performance objective of the production CTR retrieval model. Furthermore, our initial single-call (batch=1) deployment established that this offline AUC strictly correlates with the RIG observed in live production. But AUC is an aggregate: a
single scalar summarising how well the LR scorer separates clickers
from non-clickers given the \emph{whole} one-hot indicator matrix
at once. When two configurations of the agent land on the same
downstream AUC, or when AUC moves and we want to know
\emph{why}, that scalar hides the underlying behavior.

We propose Volume-Weighted Absolute Lift (VWAL) as a 
per-feature signal-mass companion to AUC. VWAL is \emph{not} a
replacement for AUC. It does not measure retrieval quality
directly. It is a diagnostic: a number whose role is to
help us interpret what is happening at the tag level when
AUC moves, stays flat, or disagrees with intuition between two
cells.

VWAL entails three main ideas:
\begin{description}
\item[Fractional Attribution] 
If a creative carries $k$ tags in a feature's cell, each tag
receives a $1/k$ share of that creative's view and a $1/k$ share
of its click. The sum of fractional views across all tags in a
feature equals the row count exactly --- the metric respects the
data set size and cannot inflate itself through multi-tag accounting.
\item[Bayesian Smoothing]
Per-tag click rates are smoothed toward the global base rate $\mu$
with confidence parameter $C$:
\begin{equation}
\hat{p} = \frac{c + C \cdot \mu}{v + C}
\label{eq:smooth}
\end{equation}
where $c$ is fractional clicks, and $v$ is fractional views. A tag
with one view and one click no longer reports $1.0$. The prior
squashes single-impression noise toward the base rate. We use
$C = 10$ throughout as a regularization prior, mimicking the empirical 
Bayes smoothing techniques formally established for target encoding 
in high-cardinality categorical features \cite{miccibarreca2001preprocessing}. 
Crucially, setting $C=10$ strikes the optimal bias-variance tradeoff for 
LLM-generated metadata: it is large enough to neutralize single-creative 
flukes, yet small enough that highly effective, long-tail tags can still 
overcome the prior and contribute to the overall VWAL after only a few 
dozen empirical impressions.

\item[Volume Weighting]
The per-tag contribution is $|\hat{p} - \mu| \cdot v$, summed
across the feature's tags and then across features. Conceptually mirroring 
Information Value (IV) used in logistic regression feature selection 
\cite{siddiqi2006credit}, this quantity is interpretable as 
\emph{impressions-worth-of-signal} --- ``the effective number of 
predicted impressions the feature confidently extracts 
from the base-rate noise.'' By multiplying probability shift (lift) 
by real-world fractional scale (volume), VWAL quantifies 
the actionable signal mass provided to the ranker.
\end{description}

To use VWAL as an effective diagnostic tool, we analyze it jointly with ROC-AUC by normalizing both metrics against an unbatched, single-call baseline. This anchor represents the ideal "identical-prompt-per-ad" scenario, where the agent processes one creative at a time without any batching interference (which, in our experiments, corresponds to the unbatched baseline for any given model). By plotting any batched configuration relative to this anchor, we establish a two-dimensional diagnostic space:The $x$-axis represents the relative downstream retrieval quality (AUC).The $y$-axis represents the relative per-feature signal mass (VWAL).The $(1.0, 1.0)$ coordinate represents the unbatched baseline. This intersection cleanly partitions the space into four distinct quadrants. By labeling these quadrants with a compact High/Low (H/L) code—comparing the AUC axis against the VWAL axis—we can diagnose exactly how a batching strategy alters the LLM's behavior (see Table \ref{tab:vwal_quad}).

\begin{table}[ht!]
\centering
\caption{Four Quadrants AUC vs VWAL}
\label{tab:vwal_quad}
\small
\begin{tabular}{|l|l|l|}
\hline
Position & Code & What it indicates \\
\hline
Top-Right & \textbf{H-AUC, H-VWAL} & Better discrimination and\\
& & more per-tag signal mass. \\
Bottom-Right & \textbf{H-AUC, L-VWAL} & Better discrimination from\\
& & fewer but stronger tags. \\
Top-Left & \textbf{L-AUC, H-VWAL} & More per-tag signal mass the\\
& & linear ranker cannot turn into \\
& & incremental discrimination. \\
Bottom-Left & \textbf{L-AUC, L-VWAL} & Both metrics drop together. \\
\hline
\end{tabular}
\end{table}

\section{Evaluation}
\label{sec:exp}

As PromptPack is designed to enrich data for a retrieval scorer, we adopt a downstream evaluation strategy. 
The quality of the agent's output is directly measured by its impact on the ranking performance of a downstream CTR prediction model.

The data set consists of 10,000 unique ad-creative titles balanced on the click label. This data set was created from our production data. The set was constructed so that it is aligned with the production regime. Behavior at this scale offered solid offline->online generalization.
Median title length is $\sim 12$ words. Titles are unique by content hash, so every LLM call enriches a distinct title.

Each enriched-CSV cell is a multi-tag string. We expand into a
binary one-hot indicator matrix with columns namespaced by feature
(\texttt{topic:finance}, \texttt{sentiment:positive}, etc.), filter
zero-variance columns, and fit a logistic regression classifier.
The classifier is evaluated using randomized 3-fold stratified cross-validation.
We report mean AUC across folds with the across-fold standard
deviation. This evaluation harness is the minimal working example that enabled us to end-to-end study behavior of PromptPack without losing generality.

For our evaluation we used four small/fast-tier closed LLM models.
Our evaluation panel consists of four closed-weight models: \path{gpt-4.1-nano}, \path{gpt-4o-mini} from OpenAI, \path{claude-haiku-4.5} from Anthropic, and \path{gemini-2.5-flash} from Google. All four are accessed through the same
internal proxy. 
We turned off \path{gemini-2.5-flash}'s reasoning capabilities  to ensure the same inference behaviour across all LLMs.
On every model, each method is evaluated across five in-context batch sizes: $\{1, 2, 5, 10, 20\}$. The five methods are designed to isolate distinct levers ---
concurrency, prompt structure, in-context batching, and
permutation-based ensembling --- so the contribution of each lever
to the downstream-retrieval AUC and the per-item cost can be
attributed independently.

\begin{description}
\item[(a) No-batching baseline.] PromptPack at
$\mathit{bs}{=}1$: one creative per LLM call, but the call still
carries the full PromptPack system prompt (taxonomy, format rules,
confidence rubric, the five per-feature definitions). This is the
\emph{quality ceiling} for any batched variant on the same prompt.
If a batched method matches this AUC, the batching is quality-preserving. 

\item[(b) PromptPack (XML).] The main method. Identical system
prompt to (a); the user prompt wraps the batch in the XML envelope
described in Section~\ref{sec:method-xml}; output is JSON aligned
by \texttt{row\_id}; the correction layer
(Section~\ref{sec:method-correction}) runs end-to-end. This configuration we considered for production.

\item[(c) PromptPack (no XML).] Drop-in replacement for (b) that
substitutes the XML envelope with a zero-indexed numbered-list
delimiter (\texttt{"0. Title: ...\textbackslash n1. Title: ..."}).
This ablation isolates the contribution of the structural envelope from the
contribution of the system prompt and the
correction layer. It directly answers the question
``how much of PromptPack's quality is the XML?''.

\item[(d) BatchLLM~\cite{zheng2024batchllm}.] API-side simulation
of the engine-level BatchLLM technique. The published BatchLLM is
a serving-stack optimisation that shares a common prefix across
many concurrent requests via prefix-cached attention. Because we cannot
modify the closed APIs we test against, we simulate the same
\emph{cost-and-throughput profile} at the API layer: each call
carries one item (\texttt{in\_context\_batch\_size{=}1}) with the
full PromptPack system prompt, and the labelled $\mathit{bs}$ is
re-interpreted as a concurrency knob via
\texttt{direct\_concurrency = bs $\times$ 16}. We include this baseline 
for two critical reasons. First, it allows us to benchmark the relative 
enrichment performance of our straightforward approach against a theoretically 
optimal, highly parallelized LLM architecture. Second, it establishes a 
rigorous baseline to verify downstream AUC stability. Although generating 
with a temperature of zero should theoretically yield fully deterministic outputs, 
real-world closed APIs occasionally exhibit slight variations across highly 
concurrent requests. This baseline helps us isolate whether any observed 
AUC variance is caused by our sequence packing or simply inherent API fluctuation.

\item[(e) BatchPrompt + BPE~\cite{lin2024batchprompt}.] BatchPrompt
is the literature batching method. BPE (Batch Permutation
Ensembling) is its quality-recovery mechanism --- run the same
batched call $K$ times with permuted item orders, then aggregate
the per-item predictions across the $K$ rounds. The published BPE
rule is single-label majority voting. our outputs are multi-label
tag sets per feature, so we adapt the rule to per-tag
multi-label majority voting: a tag is retained for a given
(item, feature) cell iff it appears in $> K/2$ of the $K$ rounds.
We sweep $K \in \{3, 5\}$ and run both the XML and no-XML
variants. 
Because the $K \times$ multiplicative API cost makes processing the full $N{=}10{,}000$ data set financially infeasible across all permutations, BPE cells are evaluated on a representative subset of $N{=}500$. 
The resulting efficiency metrics are subsequently scaled to 10,000 samples to provide a comparable projection of production throughput and cost.
\end{description}
\section{Results and Discussion}
\label{sec:results}

\begin{figure*}[h]
\centering
\includegraphics[width=\textwidth]{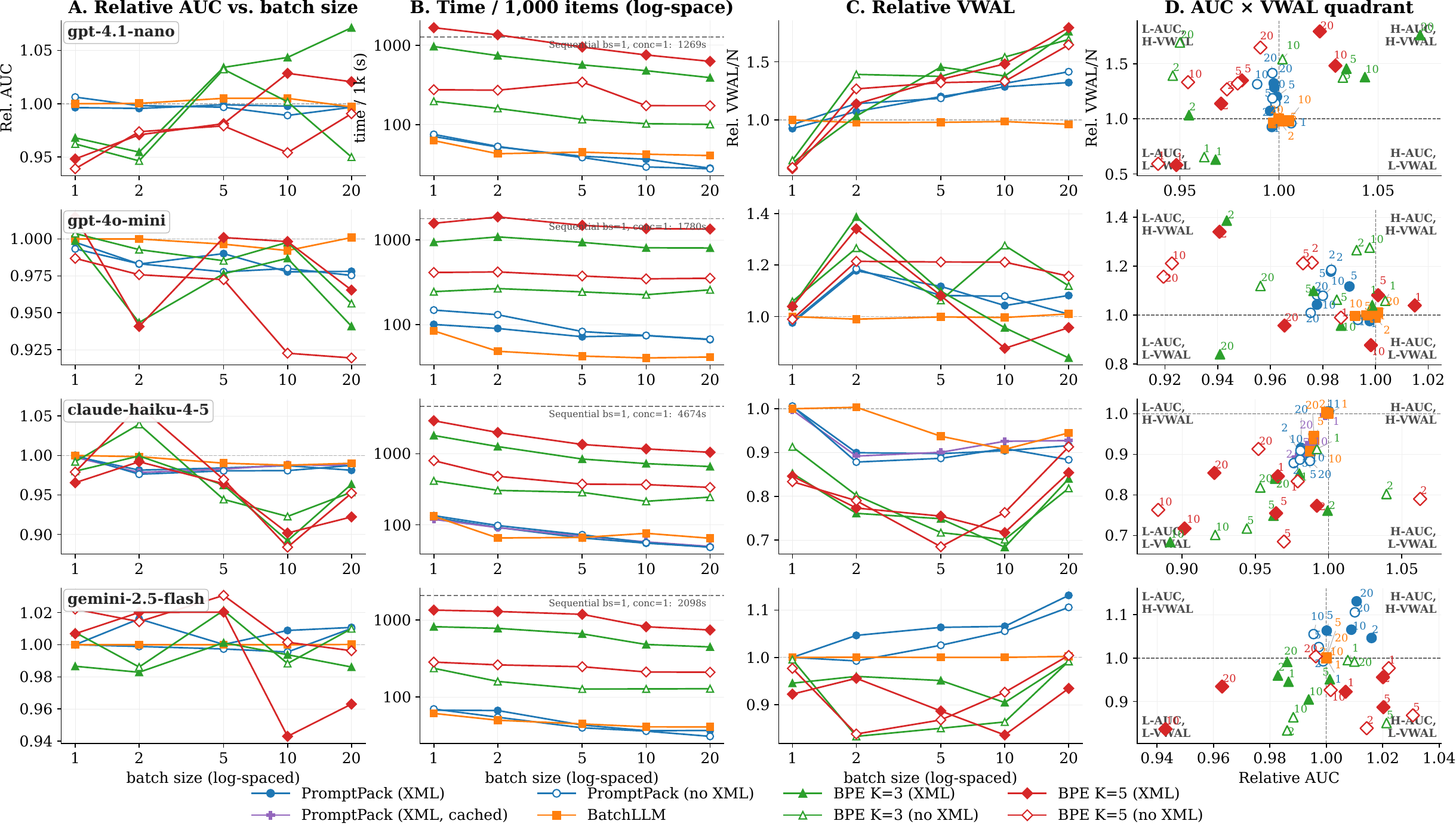}
\caption{Per-model results across the full (method, batch-size)
sweep. Rows are the four LLMs; columns are downstream-retrieval
ROC-AUC, enrichment-process time per 1k items, relative VWAL, and
the AUC$\times$VWAL quadrant view anchored at BatchLLM bs=1.}
\label{fig:per-model-4panel}
\end{figure*}

We report results on three axes per (model, method, batch\_size)
cell: the agent's downstream-retrieval ROC-AUC, the agent's
enrichment-process time per one thousand annotated creatives, and
relative VWAL. Figure~\ref{fig:per-model-4panel} shows the full
sweep at a glance. The rest of this section is organised in three
movements: AUC first (\ref{sec:results-auc}), then throughput
and cost (\ref{sec:results-time}), then the joint AUC$\,\times\,$VWAL
diagnostic that explains \emph{why} the AUC curves behave the way
they do (\ref{sec:results-vwal}). Table \ref{tab:auc-headline} summarizes the downstream-retrieval performance and confidence intervals for four LLMs across varying batch sizes and prompt formulations.

\subsection{Retrieval Quality}
\label{sec:results-auc}

Our primary engineering objective was to successfully scale the low-cost model \texttt{gpt-4.1-nano} to a high-throughput batch size of $bs=20$ without compromising downstream predictive utility. As demonstrated in Table~\ref{tab:auc-headline}, this goal was achieved: PromptPack (XML) at $bs=20$ maintains an identical downstream retrieval ROC-AUC compared to its single-call baseline ($0.609$ vs. $0.608$), while successfully cutting wall-clock processing time from 71 to 28 seconds per 1,000 creatives. Crucially, this robust behavior generalizes seamlessly across the larger, commercial models we tested. Across the entire model suite, the absolute worst-case performance degradation observed at $bs=20$ is a negligible $0.012$ AUC points (in the case of \texttt{gpt-4o-mini}). This marginal variance sits well within the baseline's 95\% confidence interval ($\pm 0.039$), proving that PromptPack's markup-guided sequencing successfully preserves critical feature signals at scale across diverse model architectures.

A second observation on the  \texttt{gpt-4.1-nano} model is that
permutation ensembling actively \emph{improves} AUC over the bs=1
baseline. BPE at $K{=}3$ XML bs=20 produces $0.654$ (the highest
single AUC the model reaches anywhere in the sweep), an absolute
$+0.043$ over plain BatchLLM bs=1. Through majority-voting across permuted rounds, the system aggressively filters out the high-variance noise inherent to the small model, effectively reducing the total volume of generated tags via pruning. This mechanism ensures that only a concentrated subset of high-confidence, strong-signal features is retained, which the downstream linear ranker can then efficiently exploit. 
The catch is the $K \times$ multiplicative cost. 
We revisit that trade-off in Section~\ref{sec:results-time}.
 
The two larger models in the panel sit at meaningfully higher absolute AUC than the production
\texttt{gpt-4.1-nano} cell across the whole sweep. \texttt{claude-haiku-4.5} is the strongest model on AUC
overall. It lands at $0.656$ at bs=1 and $0.645$ at bs=20 with
PromptPack XML. \texttt{gemini-2.5-flash} emerges as the most batching-neutral model in the panel when comparing $bs=1$ to $bs=20$. Its AUC remains remarkably stable across batch sizes and envelope variants, sitting flat or exhibiting a slight upward trend from $0.641$ to $0.648$. Much like \texttt{claude-haiku-4.5}, it comfortably clears the ultra-low-cost \texttt{gpt-4.1-nano} baseline by several AUC points across the board. Crucially, at a batch size of twenty, our standard PromptPack approach ($0.648$) strictly outperforms the more complex BPE configurations, which degrade to $0.633$ ($K=3$) and $0.618$ ($K=5$). From an operational standpoint, this batching neutrality provides an incredibly valuable production insight: although Gemini carries a higher baseline token price than the nano model, the ability to completely bypass the $K\times$ cost multiplier of BPE fundamentally shifts the deployment economics. 

The fourth model in the panel is the most batching-sensitive: AUC
drops by about one percentage point between bs=1 and bs=20
($\sim 0.623 \to 0.611$) --- roughly twice the cross-validation
fold-std and therefore a real degradation, not noise. Combined
with its higher per-token list price compared to the nano model and its lack of an absolute-AUC advantage over the
other two larger models, \texttt{gpt-4o-mini} is dominated on both
axes: it is neither the cheapest model nor the most accurate. We
include it for completeness of the panel but do not consider it a
deployment candidate.
\subsection{Throughput and Cost}
\label{sec:results-time}

Table~\ref{tab:speedup} shows the enrichment processing time for $1{,}000$ PromptPack creatives at batch sizes $1$ and $20$.

\begin{table}[ht!]
\centering
\caption{In-context Batching Speedup}
\label{tab:speedup}
\small
\setlength{\tabcolsep}{3pt}
\begin{tabular}{|l|c|c|c|}
\hline
Model & $t/\text{k}$ ($bs{=}20$) & $t/\text{k}$ ($bs{=}1$) & Speedup \\
\hline
\texttt{gpt-4.1-nano}            & 28.2 s & 70.7 s  & $2.5\times$ \\
\texttt{gpt-4o-mini}             & 66.8 s & 100.8 s & $1.5\times$ \\
\texttt{claude-haiku-4.5}        & 48.7 s & 126.9 s & $2.6\times$ \\
\texttt{claude-haiku-4.5} (cache)& 50.0 s & 119.0 s & $2.4\times$ \\
\texttt{gemini-2.5-flash}        & 36.5 s & 67.2 s  & $1.8\times$ \\
\hline
\end{tabular}
\end{table}

As compiled in Table~\ref{tab:speedup}, PromptPack delivers substantial wall-clock efficiency gains across the entire model panel, yielding processing speedups between $1.5\times$ and $2.6\times$. Crucially, on our designated core production model (\texttt{gpt-4.1-nano}), the sequence-packing strategy compresses the processing time per $1{,}000$ creatives from $70.7$ seconds down to just $28.2$ seconds---a clean $2.5\times$ throughput improvement. For a high-volume ad-retrieval pipeline, this sharp reduction in wall-clock execution time is operationally transformative, drastically accelerating offline feature store enrichment cycles while minimizing API-layer latency bottlenecks.

A natural question is why $\mathit{bs}{=}20$ does not produce a $20 \times$ throughput improvement. The answer is structural and holds regardless of API throttling. Batching compresses only the \emph{request-side} cost of the LLM call: the shared feature-taxonomy system prompt (the $\sim 8{,}500$ tokens defining the agent's behaviour) is sent once per batch instead of once per ad. The \emph{response-side} cost, generating the per-item JSON record for each creative, does not compress: with $b$ items in a batch, the LLM still has to emit $b$ separate records, each roughly the same length per item as it would be at $\mathit{bs}{=}1$. Per-call output-token generation therefore scales roughly linearly with $b$.

On modern serving stacks, output-token generation is strictly the slower of the two inference phases. Unlike the prefill phase, which processes the input prompt in a single, highly parallelized forward pass---decoding is autoregressive and cannot be parallelized within a single response sequence. To empirically isolate the performance impact of API-level prompt caching, we conducted an explicit on/off ablation using \texttt{claude-haiku-4.5}, as it uniquely allows practitioners to toggle prefix caching via a request-level parameter. In contrast, GPT and Gemini models employ implicit, always-on prompt caching mechanisms that cannot be natively deactivated for controlled A/B testing without modifying the system prompt to enforce cache misses.

As shown in Table~\ref{tab:speedup}, explicitly enabling the cache on Claude yields overall wall-clock timings that are nearly identical to the uncached baseline. This occurs because prompt caching exclusively accelerates the prefill phase, significantly reducing the Time-To-First-Token (TTFT), but it provides absolutely no acceleration for the autoregressive decode phase. Consequently, the model must still sequentially generate the massive output payload (e.g., $\sim 2{,}600$ output tokens at $\mathit{bs}{=}20$), which remains the true latency bottleneck.

The primary operational motivation for PromptPack, however, is financial cost reduction. To formally quantify these savings, let $N$ be the total titles to process, batched into requests of size $b$, requiring $\lceil N/b \rceil$ total API calls. Each request comprises a fixed system prompt of $S$ tokens and $b \cdot T$ item-specific input tokens, generating $b \cdot O$ output tokens. Let $p_{ic}$, $p_i$, and $p_o$ denote the per-token prices for cached inputs, standard inputs, and outputs, respectively. If API caching is unsupported, $p_{ic} = p_i$. The token cost of a single request is therefore:
\begin{equation}
    c_{\text{req}}(b) = \underbrace{p_{ic} \cdot S + p_i \cdot b \cdot T}_{\text{input~cost}} + \underbrace{p_o \cdot b \cdot O}_{\text{output~cost}}
    \label{eq:request_cost}
\end{equation}

When employing Batch Permutation Ensembling (BPE), each batch is independently repeated $R$ times. By packing sequences in-context, the total pipeline cost heavily amortizes the fixed system prompt overhead across $b$ titles, while scaling linearly with the ensemble size $R$:
\begin{equation}
    C_{\text{BPE}}(b, R) = R \left\lceil \frac{N}{b} \right\rceil \bigl( p_{ic}S + p_i b T + p_o b O \bigr)
    \label{eq:total_cost_bpe}
\end{equation}

To evaluate the economic viability of the agent, we measure the relative cost savings $\Delta(b, R)$ strictly against the single-run, unbatched baseline (i.e., $b{=}1, R{=}1$):
\begin{equation}
    \Delta(b, R) = 1 - \frac{C_{\text{BPE}}(b, R)}{C_{\text{BPE}}(1, 1)} = 1 - \frac{R \left\lceil N/b \right\rceil \bigl( p_{ic}S + p_i b T + p_o b O \bigr)}{N \bigl( p_{ic}S + p_i T + p_o O \bigr)}
    \label{eq:savings}
\end{equation}

A positive $\Delta$ indicates a net cost reduction. While sequence packing aggressively drives $\Delta$ higher by sharing the $p_{ic}S$ burden, the ensembling multiplier $R > 1$ actively works against these savings. Therefore, practitioners utilizing multi-pass validation must guarantee the batch size $b$ is sufficiently large to maintain a positive break-even threshold ($\Delta > 0$).

Our deployed feature taxonomy carries $S \approx 8{,}500$ system-prompt tokens. By scaling from $\mathit{bs}{=}1$ to $\mathit{bs}{=}20$, PromptPack amortizes this fixed overhead to achieve a total per-item cost reduction of $\mathbf{\sim 89\%}$.

In our production pipeline, offline LLM enrichment requests are routed through the providers' asynchronous batch APIs. Prior to introducing PromptPack, we utilized this asynchronous tier to cut our token costs by $50\%$ without any loss in prediction accuracy. 

However, compounding this $50\%$ asynchronous discount with API-level prompt caching introduces model-specific constraints. For our designated core production model (\texttt{gpt-4.1-nano}), OpenAI restricts prompt caching functionality on the Asynchronous Batch API to the \texttt{gpt-5} family. Therefore, API-level caching is technically unavailable for our primary baseline. 

Conversely, the remaining foundation models in our panel do support the simultaneous use of asynchronous batching and prompt caching. For these cache-enabled models, the heavily discounted cached input tokens ( $80\%-90\%$ cheaper) yield an immediate baseline cost reduction of $70\%-80\%$ compared to the uncached asynchronous version. Crucially, when we apply PromptPack at $\mathit{bs}{=}20$ to these cache-enabled models, our in-context sequence packing amortizes the remaining cached prefix costs, delivering an \emph{additional} $52\%$-$57\%$ cost reduction over the already-cached $\mathit{bs}{=}1$ baseline.

For our current production system PromptPack at $\mathit{bs}{=}20$ on \texttt{gpt-4.1-nano}
gives the full $89\%$ cost reduction with AUC indistinguishable
from bs=1. Despite lacking prompt caching during asynchronous batching, \texttt{gpt-4.1-nano} still yields the lowest total cost among all the configurations tested. Adding $K{=}3$ BPE buys $\sim 0.05$
extra AUC on \texttt{gpt-4.1-nano} but cuts the cost saving in
roughly two-thirds ($67.1\%$); $K{=}5$ cuts it in roughly half ($45.1\%$). For a
deployment whose primary constraint is the cost gate, plain bs=20
is the right operating point. For a deployment whose primary
constraint is squeezing the last hundredths of AUC out of the
small model, $K{=}3$ is the natural next step --- but we have to evaluate if  switching to 
\texttt{gemini-2.5-flash} at plain
bs=20, which delivers a larger AUC gain makes more sense, since the 
$K \times$ cost penalty reaches for $K=5$ for \texttt{gpt-4.1-nano}  almost the same cost as running the enrichment with \texttt{gemini-2.5-flash}.

\subsection{Interpreting Retrieval Performance via AUC-VWAL Quadrants}
\label{sec:results-vwal}

The AUC story alone is incomplete. It cannot answer why \texttt{gpt-4.1-nano}'s AUC stays flat under batching while the per-tag signal mass grows, or why a small AUC drop on \texttt{claude-haiku-4.5} is real signal loss while a similar gap on \texttt{gemini-2.5-flash} is statistical noise. To resolve this, we use a joint AUC$\,\times\,$VWAL diagnostic view (Section~\ref{sec:vwal}), anchored at BatchLLM $bs=1$, mapping each configuration into one of four quadrants. 

\begin{description}
\item[High AUC, High VWAL] \hfill \\
\emph{Example: \texttt{gpt-4.1-nano} BPE $K{=}3$ (XML), $bs=20$.}
This quadrant represents a state of optimal signal density. While permutation voting effectively prunes the high-variance noise inherent to the small model, the large batch size simultaneously ensures a high volume of committed tags per creative. As a result, the downstream linear ranker is supplied with a dense, high-purity feature distribution, maximizing predictive accuracy (AUC) and total systemic signal mass (VWAL).

\item[High AUC, Low VWAL] \hfill \\
\emph{Example: \texttt{gemini-2.5-flash} BPE $K{=}5$ (no XML), $bs=1$}.
This state is driven by strict voting. Because a tag must appear across multiple passes to survive, the system throws away a lot of marginal tags. This leaves far fewer tags per ad, which drives down VWAL. However, the tags that do survive are highly accurate. This gives the ranker a sparse but very clean input, raising the AUC. Importantly, the LLM itself isn't generating less text. Our data shows empty-cell rates actually drop under batching. The low volume is entirely caused by the strict voting filter.

\item[Flat/Low AUC, High VWAL] \hfill \\
\emph{Example: \texttt{gpt-4.1-nano} PromptPack (XML), $bs=20$.} 
This is the most populated off-anchor quadrant. The model emits more confident, voluminous tag distributions under batching (raising VWAL by up to $43\%$), but AUC remains flat. A follow-up L1-regularization test indicates that collinear redundancy (e.g., emitting \texttt{technology} alongside \texttt{smart-home}) accounts for only $\approx 10\%$ of this gap. The remainder suggests the linear ranker has simply saturated on available co-occurrence patterns, indicating a richer downstream non-linear ranker might be required to utilize the extra signal.

\item[Low AUC, Low VWAL] \hfill \\
\emph{Example: \texttt{claude-haiku-4.5} plain PromptPack, $bs=20$.} 
When both metrics drop, the loss is genuine per-tag signal degradation, not evaluation noise. This occurs either when batching genuinely degrades a larger model's outputs, or when a restrictive $K{=}5$ vote strips marginal tags from an already-sparse, unbatched ($bs=1$) output. These represent clear deployment warnings.
\end{description}

\subsection{Component Analysis: Formatting and Ensembling}
To isolate the effect of our markup design, we replaced the XML envelope with a standard numbered list, keeping all other prompt instructions identical. At $N{=}10{,}000$, the downstream predictive quality of both formats is nearly identical across the three larger models, with any AUC gaps falling strictly within the cross-validation standard deviation. The exception is \texttt{gpt-4.1-nano}. While XML and no-XML perform identically under our PromptPack configuration (both scoring $0.609$ at $bs=20$), the XML structure proves absolutely critical for stabilizing the smaller model under complex multi-pass ensembling. For instance, applying BPE $K{=}3$ without XML at $bs=20$ degrades the model's AUC to $0.580$. However, adding the XML envelope restores it to a peak of $0.654$.

Crucially, the XML envelope provides substantial operational value that AUC alone cannot measure. It guarantees delimiter-collision robustness, utilizes explicit \texttt{row\_id} tags so downstream parsers can easily recover from out-of-order or partial API responses, enables strict schema invariant checks and ensures forward compatibility for richer per-item payloads. While highly capable models do not strictly require XML to maintain accuracy, the envelope remains a component for robustness in production.

Next, we evaluated whether the computational and financial overhead of BPE permutation voting is justified. On the three larger models (\texttt{gpt-4o-mini}, \texttt{claude-haiku-4.5}, \texttt{gemini-2.5-flash}), BPE provides no consistent AUC advantage over standard PromptPack at any permutation depth ($K$) or batch size. For \texttt{gpt-4.1-nano}, BPE at $K{=}3$ (XML) and $bs=20$ does yield a statistically significant peak AUC of $0.654$. However, this gain incurs a huge wall-clock penalty. Achieving this peak takes nearly $14\times$ longer to process ($390$ secs vs $28$ secs per 1k creatives) alongside $3\times$ costs.

\begin{table*}[t]
\centering
\caption{Downstream metrics across four LLMs at $bs{=}1$ and $bs{=}20$. \textbf{AUC $\pm$ 95\% CI}: Across-fold mean with $t$-based confidence intervals. \textbf{$t/\text{k}$}: Wall-clock seconds per $1{,}000$ creatives. \textbf{VWAL}: Total Volume-Weighted Absolute Lift. \textbf{\#tags}: Distinct (feature, tag) pairs emitted. \textit{Note:} To ensure direct comparability with the $N{=}10{,}000$ baseline, BPE metrics (measured at $N{=}500$) are linearly rescaled for VWAL, and AUC CIs are adjusted for sample-size variance. The \texttt{claude-haiku-4-5} \textbf{(cached)} row confirms that native prompt caching performs within run-to-run noise of the uncached baseline.}
\label{tab:auc-headline}
\scriptsize
\setlength{\tabcolsep}{2pt}
\renewcommand{\arraystretch}{1.05}
\newcommand{\auchead}{%
\toprule
& \multicolumn{4}{c}{bs = 1} & \multicolumn{4}{c}{bs = 20} \\
\cmidrule(lr){2-5}\cmidrule(lr){6-9}
Method & AUC $\pm$ CI & $t/\text{k}$ & VWAL & \#tags & AUC $\pm$ CI & $t/\text{k}$ & VWAL & \#tags \\
\midrule}
\begin{minipage}[t]{0.49\textwidth}
\centering
\textbf{\texttt{gpt-4.1-nano}}\\[2pt]
\begin{tabular}{l cccc cccc}
\auchead
PromptPack (XML)     & $0.608 \pm 0.008$ & 71   & 1226 & 156 & $0.609 \pm 0.011$ & \textbf{28}  & 1749 & 249 \\
PromptPack (no XML)  & $\mathbf{0.614} \pm 0.014$ & 75   & 1270 & 161 & $0.609 \pm 0.003$ & \textbf{28}  & 1872 & 262 \\
BatchLLM             & $0.611 \pm 0.004$ & \textbf{63}  & 1323 & 154 & $0.609 \pm 0.005$ & 41  & 1274 & 167 \\
BPE $K{=}3$ (XML)    & $0.591 \pm 0.016$ & 962  &  834 &  59 & $\mathbf{0.654} \pm 0.023$ & 390 & 2329 &  78 \\
BPE $K{=}3$ (no XML) & $0.588 \pm 0.010$ & 197  &  865 &  58 & $0.580 \pm 0.015$ & 100 & 2239 &  75 \\
BPE $K{=}5$ (XML)    & $0.579 \pm 0.005$ & 1658 &  769 &  57 & $0.623 \pm 0.027$ & 625 & 2373 &  76 \\
BPE $K{=}5$ (no XML) & $0.573 \pm 0.006$ & 275  &  784 &  56 & $0.605 \pm 0.014$ & 173 & 2178 &  70 \\
\bottomrule
\end{tabular}
\end{minipage}\hfill
\begin{minipage}[t]{0.49\textwidth}
\centering
\textbf{\texttt{gpt-4o-mini}}\\[2pt]
\begin{tabular}{l cccc cccc}
\auchead
PromptPack (XML)     & $0.623 \pm 0.039$ & 101  & 1467 & 96 & $0.611 \pm 0.016$ & 67   & 1627 & 90 \\
PromptPack (no XML)  & $0.620 \pm 0.025$ & 149  & 1474 & 95 & $0.609 \pm 0.006$ & 67   & 1517 & 92 \\
BatchLLM             & $0.624 \pm 0.039$ & \textbf{85}   & 1503 & 97 & $\mathbf{0.625} \pm 0.033$ & \textbf{42}   & 1519 & 96 \\
BPE $K{=}3$ (XML)    & $0.624 \pm 0.019$ & 941  & 1564 & 49 & $0.587 \pm 0.003$ & 803  & 1262 & 50 \\
BPE $K{=}3$ (no XML) & $0.627 \pm 0.019$ & 246  & 1592 & 49 & $0.597 \pm 0.016$ & 258  & 1683 & 54 \\
BPE $K{=}5$ (XML)    & $\mathbf{0.634} \pm 0.014$ & 1571 & 1563 & 49 & $0.603 \pm 0.009$ & 1344 & 1439 & 49 \\
BPE $K{=}5$ (no XML) & $0.616 \pm 0.016$ & 413  & 1488 & 49 & $0.574 \pm 0.014$ & 353  & 1740 & 52 \\
\bottomrule
\end{tabular}
\end{minipage}\\[6pt]
\begin{minipage}[t]{0.49\textwidth}
\centering
\textbf{\texttt{claude-haiku-4-5}}\\[2pt]
\begin{tabular}{l cccc cccc}
\auchead
PromptPack (XML)         & $0.656 \pm 0.003$ & 127  & 2655 & 148 & $0.645 \pm 0.004$ & 49   & 2420 & 129 \\
PromptPack (XML, cached) & $0.656 \pm 0.004$ & \textbf{119}  & 2634 & 150 & $0.649 \pm 0.006$ & 50   & 2450 & 128 \\
PromptPack (no XML)      & $\mathbf{0.657} \pm 0.001$ & 135  & 2658 & 149 & $0.649 \pm 0.008$ & \textbf{48}   & 2334 & 137 \\
BatchLLM                 & $\mathbf{0.657} \pm 0.002$ & 132  & 2642 & 152 & $\mathbf{0.651} \pm 0.001$ & 65   & 2496 & 146 \\
BPE $K{=}3$ (XML)        & $0.644 \pm 0.008$ & 1809 & 2252 &  80 & $0.633 \pm 0.005$ & 660  & 2219 &  69 \\
BPE $K{=}3$ (no XML)     & $0.652 \pm 0.005$ & 416  & 2412 &  78 & $0.627 \pm 0.007$ & 245  & 2162 &  67 \\
BPE $K{=}5$ (XML)        & $0.634 \pm 0.001$ & 2914 & 2235 &  80 & $0.606 \pm 0.007$ & 1048 & 2256 &  68 \\
BPE $K{=}5$ (no XML)     & $0.643 \pm 0.006$ & 800  & 2202 &  80 & $0.626 \pm 0.008$ & 336  & 2412 &  67 \\
\bottomrule
\end{tabular}
\end{minipage}\hfill
\begin{minipage}[t]{0.49\textwidth}
\centering
\textbf{\texttt{gemini-2.5-flash}}\\[2pt]
\begin{tabular}{l cccc cccc}
\auchead
PromptPack (XML)     & $0.641 \pm 0.003$ & 67   & 2284 & 76 & $\mathbf{0.648} \pm 0.015$ & 37  & 2583 & 82 \\
PromptPack (no XML)  & $0.641 \pm 0.003$ & 70   & 2284 & 76 & $\mathbf{0.648} \pm 0.009$ & \textbf{31}  & 2525 & 90 \\
BatchLLM             & $0.641 \pm 0.003$ & \textbf{61}   & 2284 & 76 & $0.642 \pm 0.003$ & 41  & 2289 & 76 \\
BPE $K{=}3$ (XML)    & $0.633 \pm 0.021$ & 821  & 2159 & 56 & $0.633 \pm 0.021$ & 447 & 2264 & 58 \\
BPE $K{=}3$ (no XML) & $0.646 \pm 0.014$ & 236  & 2273 & 55 & $\mathbf{0.648} \pm 0.009$ & 128 & 2265 & 58 \\
BPE $K{=}5$ (XML)    & $0.646 \pm 0.030$ & 1350 & 2107 & 56 & $0.618 \pm 0.019$ & 743 & 2134 & 58 \\
BPE $K{=}5$ (no XML) & $\mathbf{0.656} \pm 0.030$ & 284  & 2231 & 56 & $0.639 \pm 0.020$ & 209 & 2293 & 60 \\
\bottomrule
\end{tabular}
\end{minipage}
\end{table*}
\section{Conclusion}
\label{sec:conclusion}

Our motivation was strictly operational: reducing the cost of a successful production system. We currently run a single-call agent which utilizes \texttt{gpt-4.1-nano} to improve ad retrieval, but paying the prompt cost for every single creative is prohibitively expensive. We set out to build a batching solution to scale down those costs without introducing the heavy preprocessing or infrastructure changes typical of other frameworks. The result is our core technical contribution, PromptPack: a lightweight, markup-guided sequence packing strategy deployed directly via standard APIs. It combines an engineered system prompt that pays the taxonomy token cost once per batch, an XML structural envelope that guarantees unambiguous per-item boundaries, and a correction layer that turns best-effort JSON into deterministic feature rows.

By implementing this architecture, PromptPack met our cost-reduction goals entirely. Running on \texttt{gpt-4.1-nano} at batch size 20, the agent fully preserves the baseline AUC while cutting token costs by 89\% and running $2.5\times$ faster. Looking ahead, both \texttt{claude-haiku-4.5} and \texttt{gemini-2.5-flash} delivered the highest overall accuracy in our panel using this exact same PromptPack format---proving they do not require expensive, multi-pass BPE ensembling. Beating the nano baseline by 2 to 4 AUC points makes these models strong candidates for a future upgrade. The natural next step is a live A/B test to determine if their higher API price is justified by the retrieval-quality gain at deployment scale.

Finally, introducing Volume-Weighted Absolute Lift (VWAL) alongside AUC gave us deeper insights into the tag generation process. It allowed us to move beyond top-line accuracy and understand exactly why the models behave the way they do under different batching conditions. Because it provides such a clear read on the underlying feature density, we plan to use VWAL as a standard monitoring metric to continuously observe the agent's generative quality and stability in live production.

\bibliographystyle{ACM-Reference-Format}
\newpage
\bibliography{references}

@inproceedings{cheng2023batchprompting,
    title = "Batch Prompting: Efficient Inference with Large Language Model {API}s",
    author = "Cheng, Zhoujun  and
      Kasai, Jungo  and
      Yu, Tao",
    editor = "Wang, Mingxuan  and
      Zitouni, Imed",
    booktitle = "Proceedings of the 2023 Conference on Empirical Methods in Natural Language Processing: Industry Track",
    month = dec,
    year = "2023",
    address = "Singapore",
    publisher = "Association for Computational Linguistics",
    url = "https://aclanthology.org/2023.emnlp-industry.74/",
    doi = "10.18653/v1/2023.emnlp-industry.74",
    pages = "792--810"
}

@inproceedings{lin2024batchprompt,
 author = {Lin, Jianzhe and Diesendruck, Maurice and Du, Liang and Abraham, Robin},
 booktitle = {International Conference on Learning Representations},
 editor = {B. Kim and Y. Yue and S. Chaudhuri and K. Fragkiadaki and M. Khan and Y. Sun},
 pages = {21590--21612},
 title = {BatchPrompt: Accomplish more with less},
 url = {https://proceedings.iclr.cc/paper_files/paper/2024/file/5d8c01de2dc698c54201c1c7d0b86974-Paper-Conference.pdf},
 volume = {2024},
 year = {2024}
}

@article{cliqueparcel2024,
  author  = {Liu, Jiayi and Yang, Tinghan and Neville, Jennifer},
  title   = {{CliqueParcel}: An Approach for Batching {LLM} Prompts that Jointly Optimizes Efficiency and Faithfulness},
  journal = {arXiv preprint arXiv:2402.14833},
  year    = {2024}
}

@inproceedings{zheng2024batchllm,
  title     = {BatchLLM: Optimizing Large Batched LLM Inference with Global Prefix Sharing and Throughput-oriented Token Batching},
  author    = {Zheng, Zhen and Ji, Xin and Fang, Taosong and Zhou, F. and Liu, Chuanjie and Peng, Gang},
  booktitle = {Proceedings of Machine Learning and Systems},
  year      = {2026}
}

@article{ji2025optimized,
  author    = {Ji, Zhaoxuan and Wang, Xinlu and Luo, Zhaojing and Xie, Zhongle and Zhang, Meihui},
  title     = {Optimized Batch Prompting for Cost-effective {LLMs}},
  journal   = {Proceedings of the VLDB Endowment},
  volume    = {18},
  number    = {7},
  pages     = {2172--2184},
  year      = {2025},
  publisher = {VLDB Endowment},
  doi       = {10.14778/3734839.3734853}
}

@article{chen2024frugalgpt,
  title={FrugalGPT: How to Use Large Language Models While Reducing Cost and Improving Performance},
  author={Lingjiao Chen and Matei A. Zaharia and James Y. Zou},
  journal = {arXiv preprint arXiv:2305.05176},
  year={2023},
}

@inproceedings{broder2008computational,
  author = {Broder, Andrei Z.},
title = {Computational advertising and recommender systems},
year = {2008},
isbn = {9781605580937},
publisher = {Association for Computing Machinery},
address = {New York, NY, USA},
doi = {10.1145/1454008.1454009},
booktitle = {Proceedings of the 2008 ACM Conference on Recommender Systems},
pages = {1–2},
numpages = {2},
location = {Lausanne, Switzerland},
series = {RecSys '08}
}

@article{dave2014computational,
  author    = {Dave, Kushal and Varma, Vasudeva},
  title     = {Computational Advertising: Techniques for Targeting Relevant Ads},
  journal   = {Foundations and Trends in Information Retrieval},
  volume    = {8},
  number    = {4-5},
  pages     = {263--418},
  year      = {2014},
  doi = {10.1561/1500000045},
  publisher = {Now Publishers, Inc.}
}

@inproceedings{10.1145/3711896.3737242,
author = {Tang, Ruiming and Zhu, Chenxu and Chen, Bo and Zhang, Weipeng and Zhu, Menghui and Dai, Xinyi and Guo, Huifeng},
title = {LLM4Tag: Automatic Tagging System for Information Retrieval via Large Language Models},
year = {2025},
isbn = {9798400714542},
publisher = {Association for Computing Machinery},
address = {New York, NY, USA},
doi = {10.1145/3711896.3737242},
booktitle = {Proceedings of the 31st ACM SIGKDD Conference on Knowledge Discovery and Data Mining V.2},
pages = {4882–4890},
numpages = {9},
location = {Toronto ON, Canada},
series = {KDD '25}
}

@inproceedings{liu-etal-2025-real,
    title = "Real-time Ad Retrieval via {LLM}-generative Commercial Intention for Sponsored Search Advertising",
    author = "Liu, Tongtong  and
      Wang, Zhaohui  and
      Qin, Meiyue  and
      Lu, Zenghui  and
      Chen, Xudong  and
      Yang, Yuekui  and
      Shu, Peng",
    editor = "Christodoulopoulos, Christos  and
      Chakraborty, Tanmoy  and
      Rose, Carolyn  and
      Peng, Violet",
    booktitle = "Proceedings of the 2025 Conference on Empirical Methods in Natural Language Processing",
    month = nov,
    year = "2025",
    address = "Suzhou, China",
    publisher = "Association for Computational Linguistics",
    url = "https://aclanthology.org/2025.emnlp-main.1473/",
    doi = "10.18653/v1/2025.emnlp-main.1473",
    pages = "28948--28960",
    ISBN = "979-8-89176-332-6"
}

@article{xmlprompting2025,
  author  = {Alpay, Faruk and Alpay, Taylan},
  title   = {{XML} Prompting as Grammar-Constrained Interaction: Fixed-Point Semantics, Convergence Guarantees, and Human-{AI} Protocols},
  journal = {arXiv preprint arXiv:2509.08182},
  year    = {2025}
}

@misc{anthropic2024xml,
  author       = {{Anthropic}},
  title        = {Prompt Engineering Interactive Tutorial: Use {XML} Tags},
  year         = {2024},
  howpublished = {\url{https://docs.anthropic.com/en/docs/build-with-claude/prompt-engineering/use-xml-tags}},
  note         = {Accessed: 2026-06-21}
}

@article{miccibarreca2001preprocessing,
  author    = {Micci-Barreca, Daniele},
  title     = {A Preprocessing Scheme for High-Cardinality Categorical Attributes in Classification and Prediction Problems},
  journal   = {ACM SIGKDD Explorations Newsletter},
  volume    = {3},
  number    = {1},
  pages     = {27--32},
  year      = {2001},
  publisher = {ACM New York, NY, USA},
  doi = {10.1145/507533.507538}
}

@book{siddiqi2006credit,
  author    = {Siddiqi, Naeem},
  title     = {Credit Risk Scorecards: Developing and Implementing Intelligent Credit Scoring},
  year      = {2006},
  publisher = {John Wiley \& Sons},
  doi = {10.1002/9781119201731}
}

@InProceedings{agent0,
author="{\v{S}}krlj, Bla{\v{z}}
and Guilleminot, Beno{\^i}t
and Tori, Andra{\v{z}}",
editor="Wen, Qingsong
and Zhang, Yongfeng
and Liu, Zhiwei
and McAuley, Julian
and Wei, Hua
and Pang, Linsey
and Liu, Wei
and Yu, Philip S.",
title="Agent0: Leveraging LLM Agents to Discover Multi-value Features from Text for Enhanced Recommendations",
booktitle="AI Agent for Information Retrieval: Generating and Ranking",
year="2026",
publisher="Springer Nature Switzerland",
address="Cham",
pages="96--109",
isbn="978-3-032-11477-8"
}


\end{document}